\pdfoutput=1
\documentclass[sigconf]{acmart}

\usepackage{subcaption}
\usepackage{multirow}
\usepackage{algorithm}
\usepackage{algpseudocode}
\usepackage{xspace}
\usepackage{float}
\usepackage{hyperref}
\newcommand*{\eg}{\textit{e}.\textit{g}.\@\xspace}
\newcommand*{\ie}{\textit{i}.\textit{e}.\@\xspace}
\makeatletter
\newcommand*{\etc}{%
	\@ifnextchar{.}%
	{\textit{etc}}%
	{\textit{etc}.\@\xspace}%
}
\newcommand*{\etal}{%
	\@ifnextchar{.}%
	{\textit{et al}}%
	{\textit{et al}.\@\xspace}%
}
\makeatother

\newcommand{\figref}[1]{Figure~\ref{#1}}
\newcommand{\tabref}[1]{Table~\ref{#1}}
\newcommand{\equref}[1]{Equation~\ref{#1}}
\newcommand{\secref}[1]{Section~\ref{#1}}
\def\BibTeX{{\rm B\kern-.05em{\sc i\kern-.025em b}\kern-.08emT\kern-.1667em\lower.7ex\hbox{E}\kern-.125emX}}

\copyrightyear{2019}
\acmYear{2019}
\setcopyright{acmcopyright}
\acmConference[MM '19] {Proceedings of the 27th ACM International Conference on Multimedia}{October 21--25, 2019}{Nice, France}
\acmBooktitle{Proceedings of the 27th ACM Int'l Conf. in Multimedia (MM'19), Oct. 21--25, 2019, Nice, France}
\acmPrice{15.00}
\acmDOI{10.1145/3343031.3350909}
\acmISBN{978-1-4503-6889-6/19/10} 
\begin{document}
	
	%
	% The "title" command has an optional parameter, allowing the author to define a "short title" to be used in page headers.
	\title{Outfit Compatibility Prediction and Diagnosis with Multi-Layered Comparison Network}
	
	%
	% The "author" command and its associated commands are used to define the authors and their affiliations.
	% Of note is the shared affiliation of the first two authors, and the "authornote" and "authornotemark" commands
	% used to denote shared contribution to the research.
	\author{Xin Wang}
	\email{wangx@mail.dhu.edu.cn}
	\orcid{0000-0002-4315-1867}
	\affiliation{%
		\institution{Donghua University \&  JD AI Research}
		\city{Shanghai}
		\country{China}
	}

	\author{Bo Wu}
	\email{bo.wu@columbia.edu}
	\affiliation{%
		\institution{Columbia University \& JD AI Research}
		\city{New York}
		\country{USA}
	}

	\author{Yun Ye}
	\email{yun.ye@intel.com}
	\affiliation{%
		\institution{JD AI Research}
		\city{Beijing}
		\country{China}
	}

	\author{Yueqi Zhong}
	\email{zhyq@dhu.edu.cn}
	\affiliation{%
%		\institution{Key Laboratory of Textile Science and Technology of Ministry of Education, College of Textile, Donghua University}
		\institution{Donghua University}
		\city{Shanghai}
		\country{China}
	}

	%
	% By default, the full list of authors will be used in the page headers. Often, this list is too long, and will overlap
	% other information printed in the page headers. This command allows the author to define a more concise list
	% of authors' names for this purpose.
	\renewcommand{\shortauthors}{Xin Wang, Bo Wu, Yueqi Zhong}
	
	%
	% The abstract is a short summary of the work to be presented in the article.
	\begin{abstract}
		Existing works about fashion outfit compatibility focus on predicting the overall compatibility of a set of fashion items with their information from different modalities. However, there are few works explore how to explain the prediction, which limits the persuasiveness and effectiveness of the model. In this work, we propose an approach to not only predict but also diagnose the outfit compatibility. We introduce an end-to-end framework for this goal, which features for: (1) The overall compatibility is learned from all type-specified pairwise similarities between items, and the backpropagation gradients are used to diagnose the incompatible factors. (2) We leverage the hierarchy of CNN and compare the features at different layers to take into account the compatibilities of different aspects from the low level (such as color, texture) to the high level (such as style). To support the proposed method, we build a new type-specified outfit dataset named Polyvore-T based on Polyvore dataset. We compare our method with the prior state-of-the-art in two tasks: outfit compatibility prediction and fill-in-the-blank. Experiments show that our approach has advantages in both prediction performance and diagnosis ability.
	\end{abstract}
	
%	To overcome the limitations of a common space for representing different types of fashion items, our framework compares the feature similarity conditioned on different fashion type combinations. 

%The comparisons from different layers not only enable our model to comprehensively diagnose the outfit but also give a rich input for the prediction. 
	
	\begin{CCSXML}
		<ccs2012>
		<concept>
		<concept_id>10010147.10010178.10010224.10010225.10010231</concept_id>
		<concept_desc>Computing methodologies~Visual content-based indexing and retrieval</concept_desc>
		<concept_significance>300</concept_significance>
		</concept>
		<concept>
		<concept_id>10010147.10010257.10010293.10010294</concept_id>
		<concept_desc>Computing methodologies~Neural networks</concept_desc>
		<concept_significance>300</concept_significance>
		</concept>
		<concept>
		<concept_id>10010405.10003550.10003555</concept_id>
		<concept_desc>Applied computing~Online shopping</concept_desc>
		<concept_significance>300</concept_significance>
		</concept>
		</ccs2012>
	\end{CCSXML}
	
	\ccsdesc[300]{Computing methodologies~Visual content-based indexing and retrieval}
	\ccsdesc[300]{Computing methodologies~Neural networks}
	\ccsdesc[300]{Applied computing~Online shopping}
	
	%
	% Keywords. The author(s) should pick words that accurately describe the work being
	% presented. Separate the keywords with commas.
	\keywords{Outfit Compatibility; Fashion Recommendation; Deep Learning}
	
	%
	% This command processes the author and affiliation and title information and builds
	% the first part of the formatted document.
    \maketitle
	
%%%%%%%%% BODY TEXT

\section{Introduction}
Outfit compatibility refers to whether multiple fashion items look good if worn together. Learning outfit compatibility can help consumers compose satisfactory outfits and help enterprises analyze the market. Many works used multimedia information to answer the questions like ``Is this outfit compatible?'' \cite{li2017mining, Vasileva2018, Tangseng2018}, ``How to generate a compatible outfit?'' \cite{Han2017, Nakamura2018}. However, they lack explanations about the prediction and generation results \ie, ``Why is the outfit compatible?''. 

\begin{figure}[t]
	\begin{center}
		\includegraphics[width=0.92\linewidth]{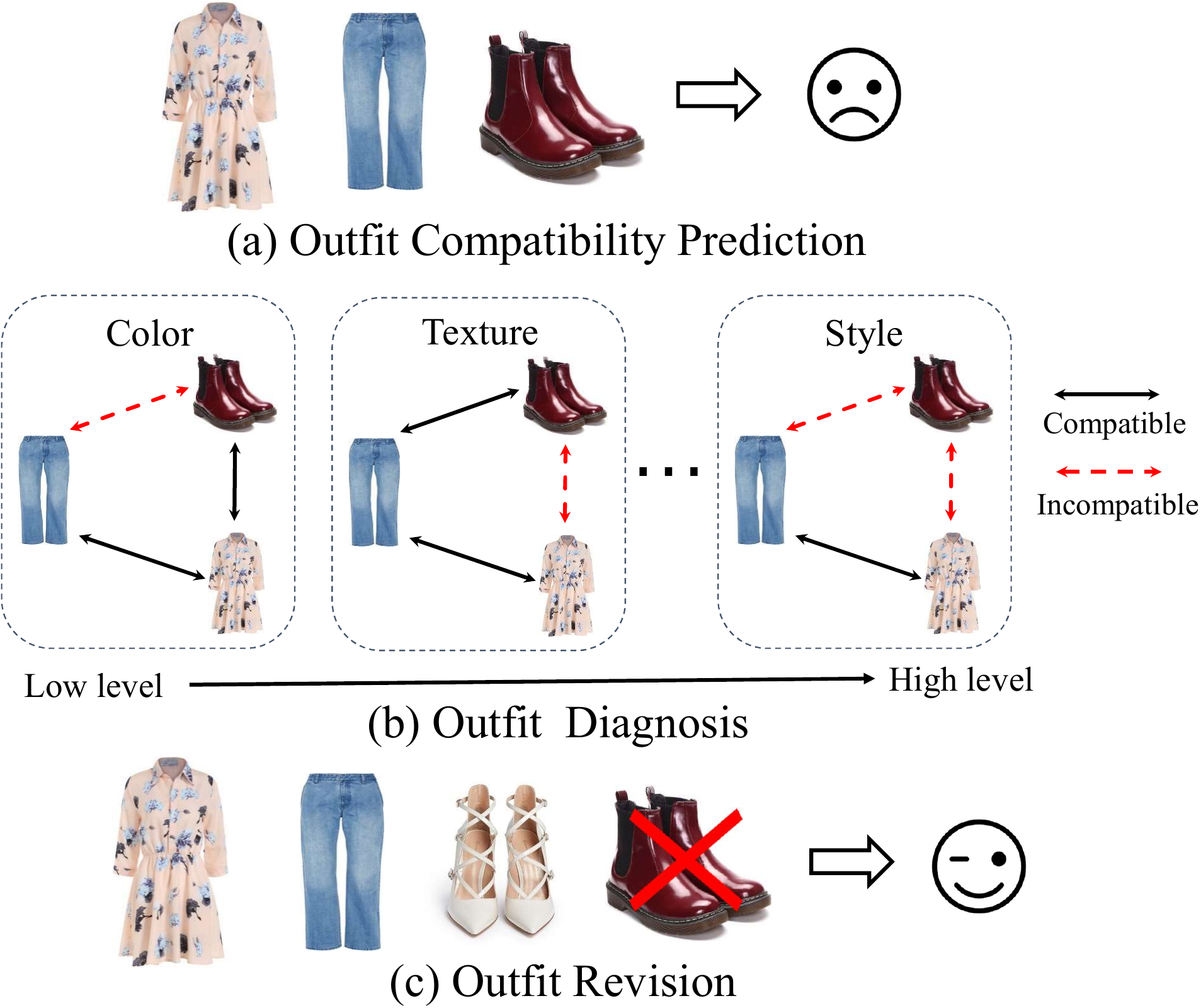}
	\end{center}
	\caption{An illustration of outfit compatibility prediction and diagnosis. (a) The prediction task tells whether the outfit is compatible. (b) The diagnosis result tells the incompatible pairs in the outfit in different aspects from the low level (such as color, texture) to the high level (such as style). The problematic pairs are marked by red dash lines. (c) Revise the outfit with little change based on the diagnosis.}
	\label{fig:task}
\end{figure}

\begin{figure*}[t]
	\begin{center}
		\includegraphics[width=0.84\linewidth]{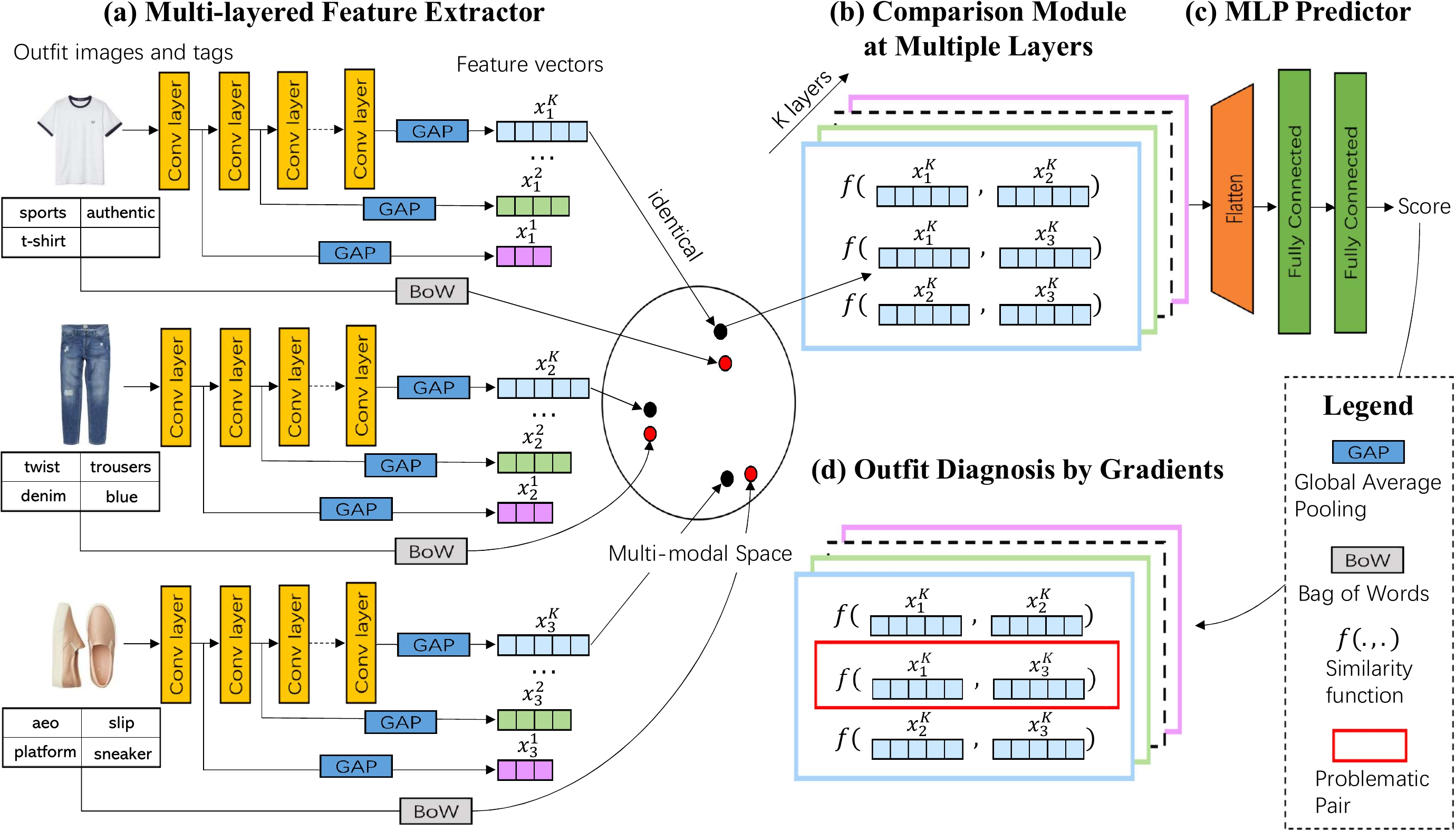}
	\end{center}
	\caption{The overview of our proposed Multi-Layered Comparison Network. The workflow of our framework is first predicting the compatibility of the outfit then use the backpropagation gradient for diagnosis. We use the feature maps at different layers of CNN and Global Average Pooling (GAP) to build representations of different aspects. The overall compatibility score is learned from the enumerated pairwise similarities at different layers. Visual Semantic Embedding (VSE) is used to handle the multimodal information of fashion items by learning a common representation between them.}
	\label{fig:overall}
\end{figure*}

\textit{Outfit diagnosis} means pointing out the incompatible factors in the fashion outfit. It provides not only the compatibility prediction but also the factors lead to incompatibility as the explanation. There are several reasons to diagnose the outfit: 1)  For researchers, it helps to understand the essence of compatibility. 2) For consumers, the explanations make the prediction and generation result more convincing. 3) For companies and designers, it provides hints for adjusting the incompatible outfit to be a more compatible and popular one. What does the diagnosis result look like? In this work, it has a format like ``the outfit is incompatible due to the incompatible color of the trousers and shoes''. As illustrated in \figref{fig:task}, Imaging a person is given an outfit for evaluation, what he or she would do is comparing each item with others and the outfit compatibility is a holistic feeling about all of the comparisons. The primitive unit in this thinking process is a single pairwise compared compatibility. A single item does not have the compatibility problem. On the other hand, fashion compatibility is evaluated in different aspects ranging from low level (such as color, texture) to high level (such as style). \cite{Zou2016} has shown that low-level features may directly determine the aesthetic feeling of fashion. 

Outfit compatibility prediction is a binary classification problem. To compute the compatibility from variable lengths of fashion items in the outfit, previous works explored the pooling operation \cite{li2017mining}, concatenation \cite{Tangseng2018} or LSTM \cite{Han2017} to consume the CNN features and output a score. Even the deeply extracted features can achieve impressive performance, it is hard to explain the meaning of each feature dimension. Different from the classification problem, the compatibility problem is irrelevant to the feature contents but lie at the comparisons between features. Besides, in previous works, only the CNN features of the last layer are used which contain strong semantic information. As mentioned before, the low-level feature is of great importance since it not only provides different aspects for diagnosis but also affects the compatibility.

\begin{table*}[!t]
	\centering
	\begin{subtable}[h]{0.6\textwidth}
		\centering
		\begin{tabular}{llccccccc}
			\toprule
			&       & Top & Bottom & Shoe  & Bag   & Accessory & Item  & Outfit \\
			\midrule
			\multirow{3}{*}{Before} & Train &   -   &    -   &   -    &   -    &   -    & 114806 & 17316 \\
			& Val &   -    &  -     &   -    &   -    &   -    & 9070  & 1497 \\
			& Test  &   -    &   -    &   -    &   -    &   -    & 18604 & 3076 \\
			\midrule
			\multirow{3}{*}{After} & Train & 13764 & 14849 & 15268 & 12640 & 12093 & 68614 & 16176 \\
			& Val & 962   & 1052  & 1124  & 948   & 823   & 4904  & 1196 \\
			& Test  & 2000  & 2153  & 2314  & 1994  & 1712  & 10173 & 2463 \\
			\bottomrule
		\end{tabular}
		\caption{The statistics before and after data cleaning}
	\end{subtable}\hfill
	\begin{subtable}[h]{0.38\textwidth}
		\centering
		\begin{tabular}{cl}
			\toprule
			Type & \multicolumn{1}{c}{Category} \\
			\midrule
			Top & Jackets, Sweaters, Coats \dots \\
			Bottom & Dresses, Skinny Jeans, Shorts \dots\\
			Shoe & Pumps, Sandals, Ankle Booties \dots \\
			Bag & Tote Bags, Clutches, Handbags \dots \\
			Accessory & Earrings, Necklaces, Sunglasses \dots \\
			\bottomrule
		\end{tabular}
		\caption{Example categories in each type group}
	\end{subtable}
	\caption{The statistics of the Polyvore-T dataset}
	\label{tab:stats}
\end{table*}

In this work, we introduce an end-to-end framework called Multi-Layered Comparison Network (MCN) to take both prediction and diagnosis into consideration. First, the overall compatibility is thought as a holistic feeling after comparing each fashion item with others, so we introduce the comparison module, which compares enumerated pairwise similarities between the input feature vectors as the output. The compatibility score is learned from all of these similarities with a non-linear model. To diagnose the outfit, we use the backpropagation gradient of each input similarity to approximate its importance for the incompatibility \cite{simonyan2013deep, springenberg2014striving}. The similarity with the largest gradient value has the most critical problem. Second, to construct features in different aspects, we leverage the hierarchical architecture of CNN, which enables the model to learn semantically stronger information as the layers going deeper. We construct the comparison modules at different layers of the backend CNN. In the early layers, the comparisons concern more about the low-level features (such as color, shape), in the later layers they concern more about the high-level features (such as category, style).

To effectively implement the comparison module, we follow the ideas from the projected embedding \cite{Chen} and type-aware embedding \cite{Vasileva2018}. To leverage both image and text information of fashion items, we adopt visual semantic embedding  \cite{Kiros2014} to learn a common representation between them. We provide a cleaner outfit dataset named Polyvore-T for the experiment, which is constructed by adding type labels and removing items that are not relevant to fashion based on \cite{Han2017}. Our approach can also be generalized to other sources of dataset. For example, to handle the fashion blogs with richer textual components, named entity recognition can be firstly used to extract the descriptions of each part in outfits, then these descriptions can be fed to our framework. We show that the diagnosis result of our method is corresponding to human cognition of the outfit compatibility. The prediction performance is tested in two fashion tasks: outfit compatibility prediction and fill-in-the-blank, in which our approach shows superiority comparing to previous works. In summary, the main contributions of this work include:

\begin{itemize}
	\item We propose to diagnose the compatibility of the outfit, which is implemented by using the gradient values to approximate the importance of input similarities.
	\item We propose to incorporate comparison modules at different layers of a convolutional neural network to account for different levels of semantic information.
	\item We build a cleaner Polyvore-T dataset for experiments and show that our framework outperforms other baselines in outfit compatibility prediction and fill-in-the-blank tasks.
\end{itemize}

%-------------------------------------------------------------------------

\section{Related Work}\label{sec:RelatedWorks}
\subsection{Fashion Recognition and Understanding}
Fashion recognition is the pioneer and basis of fashion compatibility learning. A significant amount of researches has been made to push the margin of this area in the deep learning community.  Kiapout \etal \cite{kiapour2014hipster} use a supervised method to learn 5 different fashion style with the support of human pose estimation. Liu \etal \cite{liu2016deepfashion} propose a large-scale fashion dataset called \emph{DeepFashion} and a deep network jointly training the fashion categories, attributes, and landmarks. Edgar \etal \cite{simo2016fashion} propose to learn fashion embedding using weakly-labeled data from massive online product descriptions. Lee \etal \cite{lee2017style2vec} leverage the context in fashion outfits and use a skip-gram model to learn a better fashion representation. The recognition of fashion is also affected by context information such as the city \cite{chang2017fashion}, location \cite{Zhang2017a}, scenery \cite{Simo-serra2017} and trend \cite{Lo2019}. For unsupervised learning, Hisao \etal \cite{iwata2011fashion, hsiao2017learning} use the topic model to map fashion attributes to several fashion styles by considering fashion attributes as words, outfits as documents and styles as topics. Al-Halah \etal \cite{al2017fashion} introduce a non-negative matrix factorization method to project the deeply extracted features into a fashion style space.

\subsection{Visual Compatibility Learning}

Even though compatibility is a subjective aesthetic standard, massive user feedbacks like co-view and co-purchase records \cite{McATarShiHen15, Veit} or proposals from the 	fashion community \cite{Han2017} can be used to learn the rule of compatibility. Recent works about outfit compatibility can be divided into two groups. The first group focuses on learning the pairwise compatibility, another group learns outfit compatibility in an end-to-end manner.

Pairwise compatibility can be learned as a metric by mapping item features into a compatibility space then compute their euclidean distance \cite{McATarShiHen15,Veit}. Further researches relief this metric since compatibility is not as strict as retrieval tasks. He \etal \cite{he2016learning} mix multiple metrics with different confidence to achieve more robust performance. Shih \etal \cite{Shih2017} propose to make multiple projection points of a query image. Vasileva \etal \cite{Vasileva2018} learn different metric spaces for different type combinations. While evaluating the outfit compatibility, these methods take the average of all pairwise compatibility as the output and neglect the relationship between the pairwise compatibility and the overall compatibility. 

For end-to-end training outfits, pooling \cite{li2017mining} or concatenation \cite{Tangseng2018} operations can be used to aggregate multiple item features.
Then the multilayer perceptron (MLP) can be used to compute a compatibility score. However, the addition operation in MLP is not an efficient way to learn interactions between each input \cite{Qu2017}. Han \etal \cite{Han2017} propose to use BiLSTM to construct dependency between several fashion items. Based on that, Nakamura \etal \cite{Nakamura2018} build an autoencoder upon the BiLSTM to learn the style embedding of outfits, which has similar idea as controlling the style of image captions \cite{gan2017stylenet}. However, an outfit is more like a set rather than a sequence because it is disordered.

%RNN takes in default that the neighborhood relationships like ``hat and top'' are more important than the long distance relationships like ``hat and shoes'' due to its sequence architecture.

%-------------------------------------------------------------------------

\section{Approach}\label{sec:Approach}
Our approach aims to diagnose the outfit in different aspects, which is implemented by an end-to-end framework called Multi-layered Comparison Network (MCN). There are four parts in MCN: it starts from the multi-layered feature extractor (\secref{sec:multilayer}) which extract features in different aspects. Then the comparison modules (\secref{sec:comparison}) compare enumerated pairwise similarities between features along with multiple layers and the MLP predictor compute a score from them all. Visual semantic embedding (\secref{sec:vse}) is used to handle multimodal information. The workflow of MCN is first predicting the outfit compatibility then backpropagating a gradient from the output score till the input similarities. Similar works are \cite{simonyan2013deep, springenberg2014striving}, which visualize the response of each pixel to different classes for interpreting the CNN classification, but our goal is to visualize the response of each similarity to the compatibility score. 

In the following contents of this section, we first show the diagnosis process (\secref{sec:diagnosis}) then introduce each part in the framework. The overall scheme of MCN is presented in \figref{fig:overall}.

\subsection{Outfit Diagnosis by Gradients} \label{sec:diagnosis}
Outfit compatibility can be considered as a summary after comparing each item with others in different aspects, \eg, color, texture and style \etc. To learn the relationship between the holistic compatibility and pairwise similarities, a simple linear model can be used and has great interpretability, because the weight of each input dimension indicate the importance for the output. The drawback of the linear model is the limited capacity. In contrast, multilayer perceptron (MLP) model has better capacity but the hidden unit is hard to explain. To leverage the advantage of MLP while maintaining the interpretability for diagnosis, we use the backpropagation gradients to approximate the importance of each similarity concerning the incompatibility.

Given an outfit containing $N$ items, their features in a specific aspect, \eg the color, can be denoted as a set $X=\{x_1, x_2, \dots, x_N\}$ where $x_i$ is the vector for the $i$-th item. The enumerated pairwise similarities among the features can be represented as a matrix form:

\begin{equation}
R=\left[\begin{matrix}
r_{11} & r_{12} & \cdots & r_{1N} \\
r_{21} & r_{22} & \cdots & r_{2N} \\
\vdots & \vdots & \ddots & \vdots \\
r_{N1} & r_{N2} & \cdots & r_{NN} 
\end{matrix}
\right]
\end{equation}

Here, $r_{ij}$ is the similarity between $x_i$ and $x_j$, which is an undirected relationship, formally $r_{ij} = r_{ji}$. We call $R$ the comparison matrix. For comparisons in $K$ different aspects, there are $K$ different comparison matrices $\{R^1, R^2, \dots, R^K\}$. All elements in these comparison matrices will be fed into a 2-layer MLP to compute the score of outfit compatibility:

\begin{equation} \label{eq:predictor}
s = \mathbf{W_2} ReLU(\mathbf{W_1}[R^1; R^2; \dots; R^K] + \mathbf{b})
\end{equation}

Here, $[.,.]$ denotes concatenating multiple matrices and flattening them into a vector. The non-linear relationship between $s$ and multiple $R$ can be helpful for a better prediction performance but make it hard to estimate the importance of each input to the compatibility score like a simple linear model.  To solve this, we use the derivative of $s$ to approximate the importance of each input similarities. One interpretation of this operation is if we compute the first-order Taylor expansion of the MLP model:

\begin{equation}
s \approx \mathbf{W^\dagger} [R^1; R^2; \dots; R^K] + \mathbf{b}
\end{equation}

the approximated equation is a linear model and the weight $W^\dagger$ can be easily interpreted as the importance of each input similarity. The element of $W^\dagger$ is the derivative of $s$ with respect to the point of input similarities $[R_0^1; R_0^2; \dots; R_0^K]$:

\begin{equation}
w_{ij}^k = \frac{\partial s}{\partial r_{ij}^k}\Bigr|_{[R_0^1; R_0^2; \dots; R_0^K]} .
\end{equation}

Suppose incompatible outfits are labeled as $1$, $w_{ij}^k$ can be interpreted as the importance of each similarity concerning the incompatibility (If incompatible outfits are $0$, just use the opposite number of $w_{ij}^k$). The importance of each item can be computed by summing up the gradients of all similarities containing it:

\begin{equation}
\omega_q = \sum_{1 \le k \le K} \sum_{i = q, j \neq q} w_{ij}^k
\end{equation}

where $\omega_q$ is the diagnosed importance of the $q$-th item. By substituting the most problematic item, we can revise the outfit to be a more compatible one with very little change to the original composition. During the training, we use the sigmoid function $\sigma(s) = \frac{1}{1+\exp(-s)}$ to model the output score as the compatibility probability and use the binary cross entropy as the loss function: 

\begin{equation}
\mathcal{L}_{clf} = y \cdot \log \sigma(s)
+ (1 - y) \cdot \log (1 - \sigma(s))
\end{equation}

\subsection{Comparison with Projected Embedding} \label{sec:comparison}
An important problem is how to compute each pairwise similarity $r_{ij}$ in the comparison matrix. The simplest way is adopting cosine similarity to measure the distance of features in a common space. However, a common space will induce some undesired consequences: (1) The variation of compatibility is compressed. For example, all trousers compatible with a shirt will be forced to close with each other, which means these trousers must be compatible with each other even they are not. (2) The triangle inequality will limit the position of each embedding. The triangle inequality states that for any three items, the sum of any two pairwise embedding distances should be greater than or equal to the remaining pairwise embedding distance. This will leads to an improper situation e.g., if a shirt matches a trouser, the trouser then matches a shoe, the consequence is the shoe is forced to match the shirt. Therefore, we use the projected embedding with respect to different fashion type combinations to address the above problems, which refers to \cite{Chen, Vasileva2018}.

A fashion outfit often contains items from different types such as top, bottom and shoe \etc. Different kinds of pairwise type combinations can be used as conditions for projecting the embedding to different subspaces. Let $r_{ij} = f(x_i, x_j)$ as the similarity between $x_i$ and $x_j$. We compute the similarity following a projection process:

\begin{equation}\label{eq:proj}
f(x_i, x_j) = d(P^{i \to (i,j)} x_i, P^{j \to (i,j)} x_j)
\end{equation}

Here, $P^{i \to (i,j)}$ is the projection of $i$-th item conditioned on combination $(i,j)$. $d(.,.)$ is the cosine similarity. The projection in this work is implemented as:

\begin{equation}\label{eq:mask}
P^{i \to (i,j)} x_i = ReLU(x_i \otimes m_{(i,j)})
\end{equation}

where $m_{(i,j)}$ is a learnable mask vector with the same dimension as the features $x_i$, $\otimes$ denotes the element-wise product operation. The mask $m_{(i,j)}$ works as an element-wise gating function which selects relevant elements in the feature vector for different compatibility conditions. We add two additional loss to penalize the training \cite{Veit2017}: $\mathcal{L}_{mask}$ aims to encourage the masks to be sparse and $\mathcal{L}_{emb}$ encourages the CNN to encode normalized representations in the latent space:

\begin{equation}
\mathcal{L}_{mask}(m) = ||m||_1 ,\quad \mathcal{L}_{emb}(x) = ||x||_2
\end{equation}

\begin{table}[t]
	\begin{center}
		\begin{tabular}{lcc}
			\toprule
			Method & AUC(\%) & FITB(\%) \\
			\midrule
			Pooling\cite{li2017mining} & $88.35  \pm 0.26$ & $57.28 \pm 0.31$ \\
			Concatenation\cite{Tangseng2018} & $83.40  \pm 0.48$ & $52.91 \pm 0.59$ \\
			%Siamese \cite{Veit} &  &  \\
			Self-Attention \cite{Wang} & $79.65  \pm 0.68$ & $48.60 \pm 0.70$  \\
			CSN\cite{Vasileva2018} & $84.90 \pm 0.52$ & $57.06 \pm 1.70 $ \\
			BiLSTM\cite{Han2017} & $74.44 \pm 0.95$ & $45.41 \pm 0.40$ \\
			BiLSTM+VSE\cite{Han2017}  & $74.82 \pm 0.63 $ & $46.02 \pm 0.62$ \\
			Ours & $\mathbf{91.90} \pm 0.40$ & $\mathbf{64.35} \pm 0.92$ \\
			\bottomrule
		\end{tabular}
	\end{center}
	\caption{Outfit compatibility prediction AUC and FITB accuracy on Polyvore-T dataset.}
	\label{tab:comparsion}
\end{table}

\subsection{Multi-Layered Representation} \label{sec:multilayer}
The representation of fashion items in different aspects not only enables the model to diagnose the outfit in a detailed manner but also feed rich information for compatibility prediction. To construct different representations, one way is to predefine several concepts such as color, texture and shape \etc. However, these definitions may not completely cover the fashion descriptions. Predefinition is not the only way for representations in different aspects. Modern deep learning use CNN which deeply connects multiple convolutional layers. Even though the convolutional kernel in each layer has a small size but with the architecture going deeper, the receptive field of the model can become larger. Therefore, the former layers in CNN capture low-level features such as color, texture, the later layers capture high-level features such as fashion style and compatibility. This characteristic of CNN can be used for building different representations of fashion items from low level to high level.

Leveraging this fact, we construct comparison modules at different layers of the backend CNN model. In detail, we use Global Average Pooling (GAP) to reduce the features maps at different CNN middle layers into vectors \cite{Lin2013}. GAP has two benefits in this situation. First, GAP transforms feature maps into vectors to satisfy the comparison operation in \equref{eq:proj}. Second, fashion features like color, texture \etc are irrelevant to location and GAP can effectively get rid of the spatial information. Feature vectors of $K$ layers are fed into $K$ different comparison module to compute enumerated pairwise similarities $\{R^1, R^2, \dots, R^K\}$, which is the input for computing overall compatibility in \equref{eq:predictor}. In experiment, we use ResNet-50 \cite{he2016deep} as the backend network, 4 different layers are chosen for multi-layered representation including ``conv2\_x", ``conv3\_x", ``conv4\_x", ``conv5\_x".

\subsection{Visual Semantic Embedding} \label{sec:vse}
In real-life situations, fashion items are often described by multimodal information such as image, tags \etc. Visual Semantic Embedding (VSE) \cite{Kiros2014} is a method to take full advantage of information from different modalities by providing a common expression between them\cite{Han2017,Nakamura2018}.

For one fashion item in Polyvore dataset, its associated text description is like ``classic skinny jeans'', which can be denoted as $S = \{w_1, w_2, \cdots, w_n\}$, where $w_i$ is $i$-th word and can be represented as one-hot vector $e_i$. The word embedding of $e_i$ is $v_i = W_T e_i$ and the semantic embedding of an outfit is $v = \frac{1}{M} \sum_{i=1}^M v_i$, where $W_T$ is the weight of word embedding model. A similar process is made to project the visual features $x$ into the embedding space as $u = W_I x$. 

The object of VSE is to make $u$ and $v$ close to each other in the joint space when they are from the same item. Minimizing the following contrastive loss can achieve that object:

\begin{equation}
\begin{aligned}
\mathcal{L}_{vse}(v, u;W_T, W_I) = & \sum_u \sum_k max(0, m - d(u, v) + d(u, v_k)) \\
& + \sum_v \sum_k max(0, m - d(v, u) + d(v, u_k))
\end{aligned}
\end{equation}

where $d(u, v)$ is the function to compute the distance between two embeddings. For one matching (from the same item) pair $v$ and $u$, $v_k$ denotes the semantic embeddings of all possible non-matching items, $u_k$ denotes the visual embeddings of all possible non-matching items. This loss function expects the distances of all matching pairs $u$ and $v$ are smaller than all of the other non-matching pairs with a margin $m$. In practice, we can use the mini-batch as the set to enumerate all $u_k$ and $v_k$.

To the end, our framework can be jointly trained in an end-to-end manner with a total loss $\mathcal{L}_{total}$ and $\lambda_{\{1, 2, 3\}}$ is the weights for each additional losses. The trainable parameters includes $\{\Theta$, $W_T$, $W_I$, $W_2$, $W_1$, $b$, $M\}$ where $\Theta$ is the parameters for the backend CNN model, $W_2, W_1, b$ are the parameters of MLP in \equref{eq:predictor}, $M$ denotes the masks for different conditions in \equref{eq:mask}.

\begin{equation}
\mathcal{L}_{total} = \mathcal{L}_{clf} + \lambda_1 \mathcal{L}_{emb} + \lambda_2  \mathcal{L}_{mask} + \lambda_3 \mathcal{L}_{vse}
\end{equation}

\begin{table}[t]
	\begin{center}
		\begin{tabular}{lccc}
			\toprule
			Method & AUC(\%) & FITB(\%) & \#Param\\
			\midrule
			Pooling\cite{li2017mining} & $88.35  \pm 0.26$ & $57.28 \pm 0.31$ & $ 10^6 $\\
			Concatenation\cite{Tangseng2018} & $83.40  \pm 0.48$ & $52.91 \pm 0.59$ & $ 10^6 $ \\
			CM & $87.06 \pm 1.08$ & $60.07 \pm 0.32$ & $ 10^2 $ \\
			CM + VSE & $88.52 \pm 0.94$ & $61.04 \pm 0.81$ & $10^2$ \\
			CM + PE & $90.16 \pm 0.81$ & $\mathbf{63.20} \pm 0.79$ & $ 10^3 $ \\
			CM + VSE + PE  & $\mathbf{90.70} \pm 0.54 $ & $62.47 \pm 0.19$ & $ 10^3 $ \\
			\bottomrule
		\end{tabular}
	\end{center}
	\caption{Ablation study for each part in MCN framework including the Comparison Module (CM), Visual Semantic Embedding  (VSE) and Projected Embedding (PE).}
	\label{tab:ablation1}
\end{table}

%------------------------------------------------------------------------

\section{Experiment}\label{sec:Experiment}
To overcome the subjective nature of fashion sense, a dataset containing the compatible outfits can be constructed by fashion-sensitive people. The randomly combined outfits are highly possible incompatible. In this section, we first introduce our method to build the Polyvore-T dataset. Then we compare the prediction performance of our MCN framework with several baselines. Finally, we show the result of automatically diagnosing and revising the fashion outfit. Code and dataset are released at \url{https://github.com/WangXin93/fashion_compatibility_mcn}.

\subsection{Polyvore-T Dataset}\label{subsec:Dataset}
We build a type-labeled fashion outfit dataset upon the Polyvore dataset \cite{Han2017}. Original Polyvore dataset contains 21889 expertise-selected outfits. Graph segmentation is used to split train, validation and test dataset. Each item has the corresponding image, text description, votes, and category (like jeans, skirts, sports, totally 381 kinds of category). However, there are overlaps between these categories such as ``shoulder bags`` and ``bags`` \etc. At the same time, some categories do not have insufficient samples for training. So we label type information for each item by grouping 381 categories into 5 types with the following method:  (1) Unrelated categories such as ``lipsticks'', ``mirrors'' are filtered and there are 158 categories retained after filtering. (2) The filtered categories are classified into 5 types (top, bottom, shoes, bag, accessory) by hand. (3) The item not in the category list will be dropped and the left items can be labeled according to their categories. When there are multiple items of the same type in one outfit, only the first item will be taken. We filtered out outfits in the dataset with less than 3 types to make sure each outfit in our dataset has at least 3 items and up to 5 items. The statistics of the filtered dataset are shown in \tabref{tab:stats}. Type information enables the comparisons are conditioned on the type combination as discussed in \secref{sec:comparison}. It also prevents repeating items such as two pairs of trousers in one outfit. 

\begin{figure}[t]
	\centering
	\includegraphics[width=0.98\linewidth]{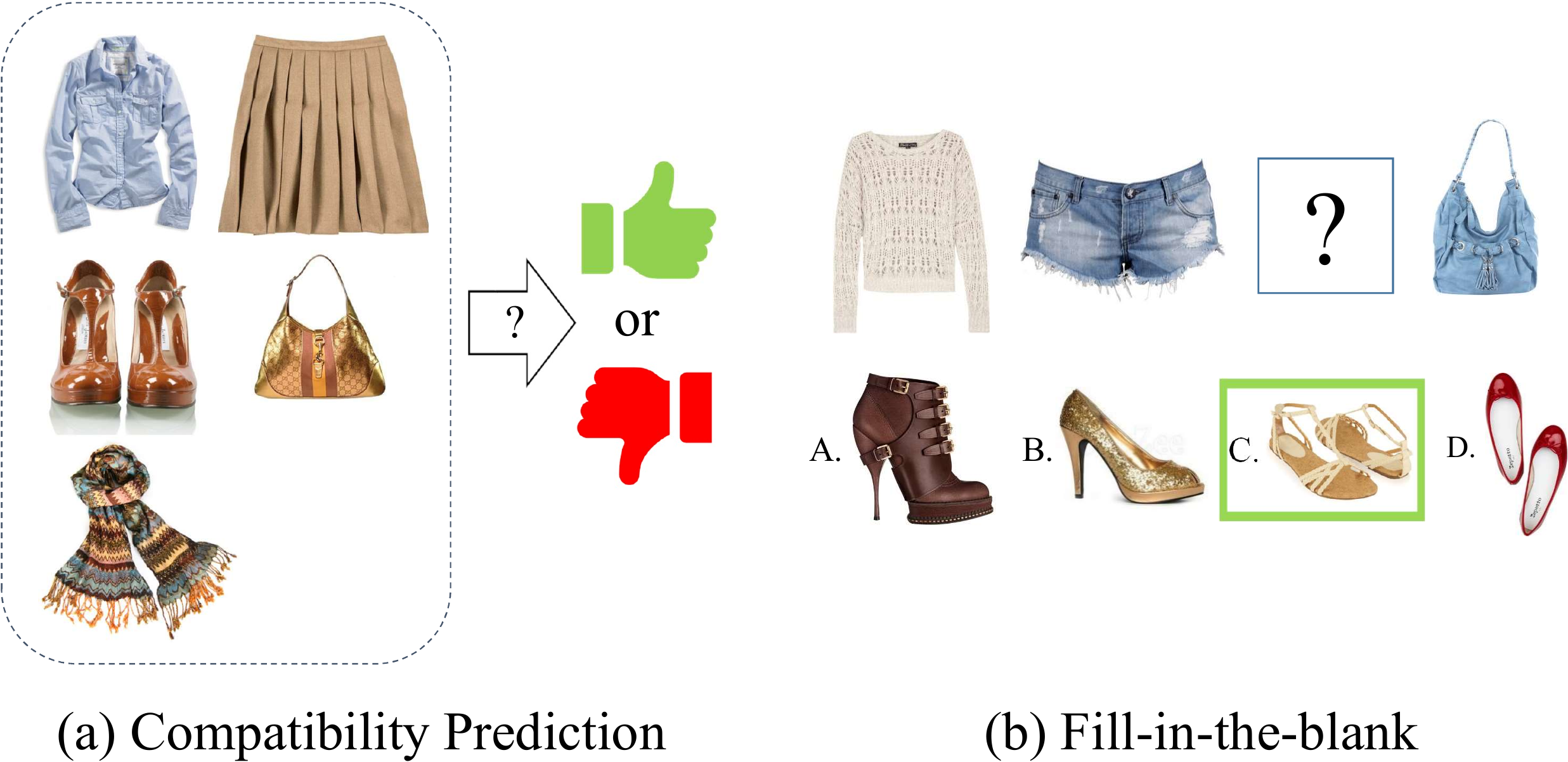}
	\caption{Two tasks for evaluating the compatibility prediction performance: (a) Given an outfit, the model predicts its compatibility. (b) Fill-in-the-blank task has a question outfit and four options.}\label{fig:eval}
\end{figure}

\subsection{Experiment Settings}\label{subsec:Details}

\textbf{Training.}
In all experiments, we use ImageNet\cite{russakovsky2015imagenet} pre-trained ResNet-50 \cite{he2016deep} model as the backbone model unless specified. The spatial size of the input images is $224 \times 224$. The input outfit has a variable length from 3 to 5.  The empty parts are handled by padding a mean image of that type. Batch normalization layer \cite{ioffe2015batch} is added after the comparison modules to normalize the similarities of the same type combination. All experiments are trained on a single Tesla P40 GPU, each input mini-batch has 32 outfits. It takes about 9 hours to train 50 epochs. The initial learning rate is $1e^{-2}$ and decay by a factor of 0.2 every 10 epochs. The optimizing strategy is SGD with a momentum of 0.9. We set the weights of additional losses $\lambda_{\{1, 2, 3\}}$ are $5e^{-3}$, $5e^{-4}$ and 1 respectively. Only model parameters with the best performance on the validation set will be saved.

\noindent\textbf{Negative Sample.}
During the training, the positive samples or compatible outfits come from the ground truth dataset. Negative samples are generated by substituting each item in the positive outfit samples with a random item of the same type from the dataset. It is because in real life fashion designers focus on composing a compatible outfit, but seldom think how to build an incompatible outfit. The subjectivity in incompatibility also makes the dataset labeling unreliable. Since fashion experts in communities like Polyvore have composed the outfits with the different aesthetic rules, randomly composed outfits are highly possible incompatible.

\subsection{Tasks} \label{subsec:tasks}

The performance of outfit compatibility prediction is evaluated in two tasks: outfit compatibility prediction AUC and fill-in-the-blank accuracy. \figref{fig:eval} gives an illustration of the two tasks. During the test, each model is evaluated 5 times with different dynamically generated test sets.

\noindent\textbf{Outfit Compatibility Prediction (AUC): }
	The goal of the outfit compatibility prediction task is to compute a score as overall compatibility. MCN predicts the score from multiple input images in an end-to-end way. We generate test set following \secref{subsec:Details} with a positive and negative ratio of $1:1$. The area under the ROC curve is used to compare different methods.
	
\noindent\textbf{Fill-in-the-blank (FITB): }
	Fill-in-the-blank task aims to select the most compatible item with the remainder of the outfit as the question. It is evaluated by the accuracy of correctly answering the question. Each question has 4 options in our experiment. MCN and other baselines do this task by predicting scores of 4 different outfits in which only substituting the blank part with different options, then choose the option with the highest score as the answer.

\begin{table}[t]
	\begin{center}
		\begin{tabular}{lcc}
			\toprule
			Method & AUC(\%) & FITB(\%) \\
			\midrule
			CSN\cite{Vasileva2018} & $84.90 \pm 0.52$ & $57.06 \pm 1.70 $ \\
			CM + Avg & $85.28 \pm 0.54$ & $55.62 \pm 0.89$ \\
			CM + Linear & $88.58 \pm 0.85$ & $60.91 \pm 1.19$ \\
			CM + 2FC  & $\mathbf{90.70} \pm 0.54 $ & $\mathbf{62.47} \pm 0.19$ \\
			CM + 3FC  & $88.53 \pm 0.60 $  & $61.40 \pm 0.75$ \\
			\bottomrule
		\end{tabular}
	\end{center}
	\caption{Results of different layers of MLP to learn overall compatibility from all pairwise similarities.}
	\label{tab:ablation2}
\end{table}

\begin{table}[t]
	\begin{center}
		\begin{tabular}{lcc}
			\toprule
			Method & AUC(\%) & FITB(\%) \\
			\midrule
			Layer 4 & $90.70 \pm 0.54 $ & $62.47 \pm 0.19$ \\
			Layer 4+3 & $90.53 \pm 0.69 $ & $64.14 \pm 1.06$ \\
			Layer 4+3+2 &  $90.92 \pm 0.53 $ & $\mathbf{64.53} \pm 0.58$ \\
			Layer 4+3+2+1 & $\mathbf{91.90} \pm 0.40$ & $64.35 \pm 0.92$ \\
			\bottomrule
		\end{tabular}
	\end{center}
	\caption{Prediction performance with representation from different layers.}
	\label{tab:scales}
\end{table}

\subsection{Baselines}\label{subsec:baselines}

We compare our method with several recent fashion outfit compatibility works. Their brief descriptions are as follows:

\noindent \textbf{Pooling \cite{li2017mining}:} 
It uses average-pooling operation to aggregate the variable length of feature inputs to predict the compatibility.
	
\noindent \textbf{Concatenation \cite{Tangseng2018}:} 
It concatenates 5 item features into a long vector with a length of $1000 \times 5$, then uses MLP as the binary classifier. We set the size of the hidden layer as 1000.
	
\noindent \textbf{CSN \cite{Vasileva2018}:}
It is a metric learning method for pairwise compatibility. The compatibility is computed as the distance of projected embeddings conditioned on different type combination. Outfit compatibility is the average of all pairwise compatibility.
	
\noindent \textbf{BiLSTM+VSE \cite{Han2017}:}
Each time step LSTM consumes one CNN encoded features and outputs a hidden state and a prediction of the next item. The cross entropy between the predicted items and the ground-truth is the compatibility score. It jointly optimizes the forward LSTM loss, backward LSTM loss, and VSE loss. 
	
\noindent \textbf{Self-Attention \cite{Wang}:}
It uses the self-attention mechanism to relate different items in an outfit to compute a representation of the outfit. We use the \textit{scaled dot-product attention} \cite{NIPS2017_7181} where the query, key, and value are the item features in the same outfit.

\begin{figure}[t]
	\centering
	\begin{subfigure}[b]{0.48\textwidth}
		\centering
		\includegraphics[width=0.9\linewidth]{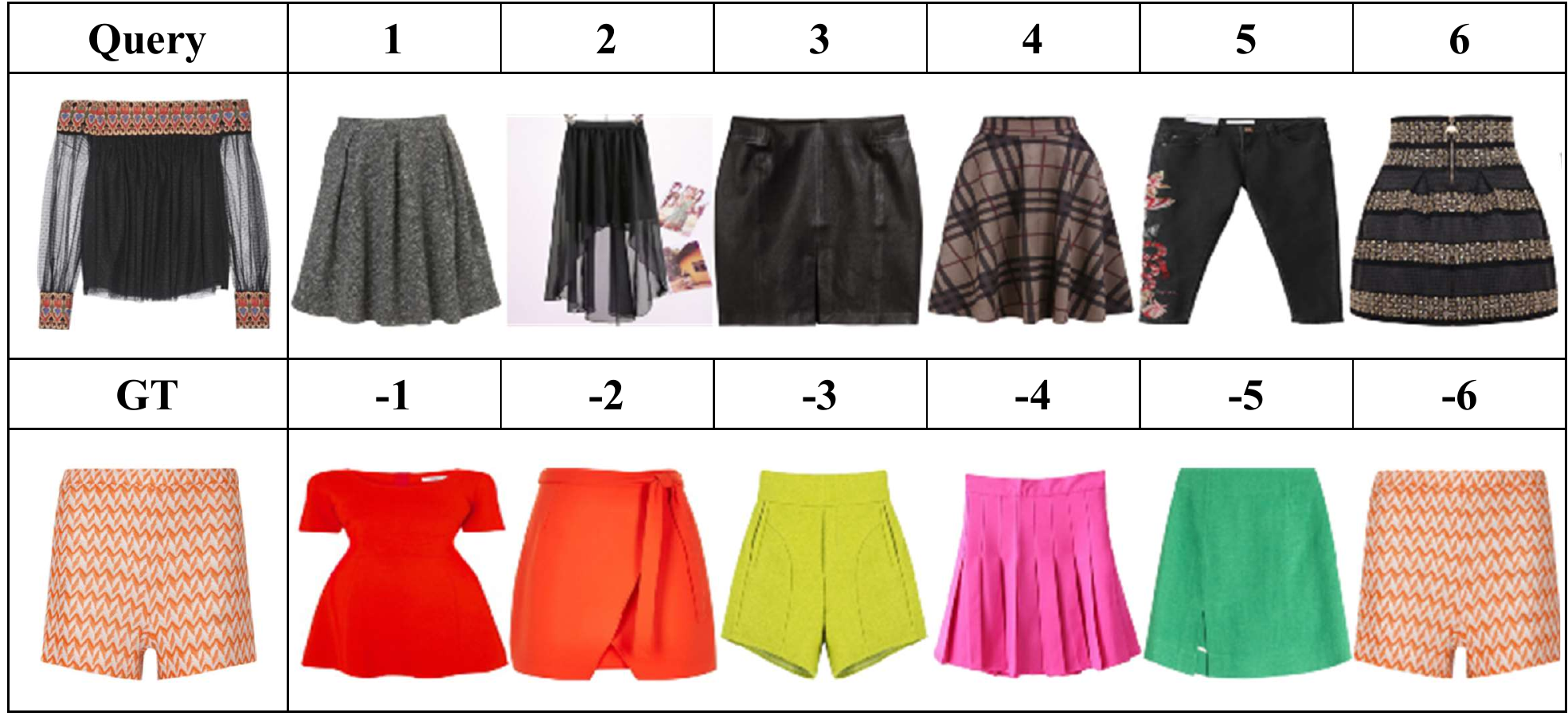}
		\caption{Retrieval results with representation at Layer 1}\label{fig:lowlevel}
	\end{subfigure}
	\begin{subfigure}[b]{0.48\textwidth}
		\centering
		\includegraphics[width=0.9\linewidth]{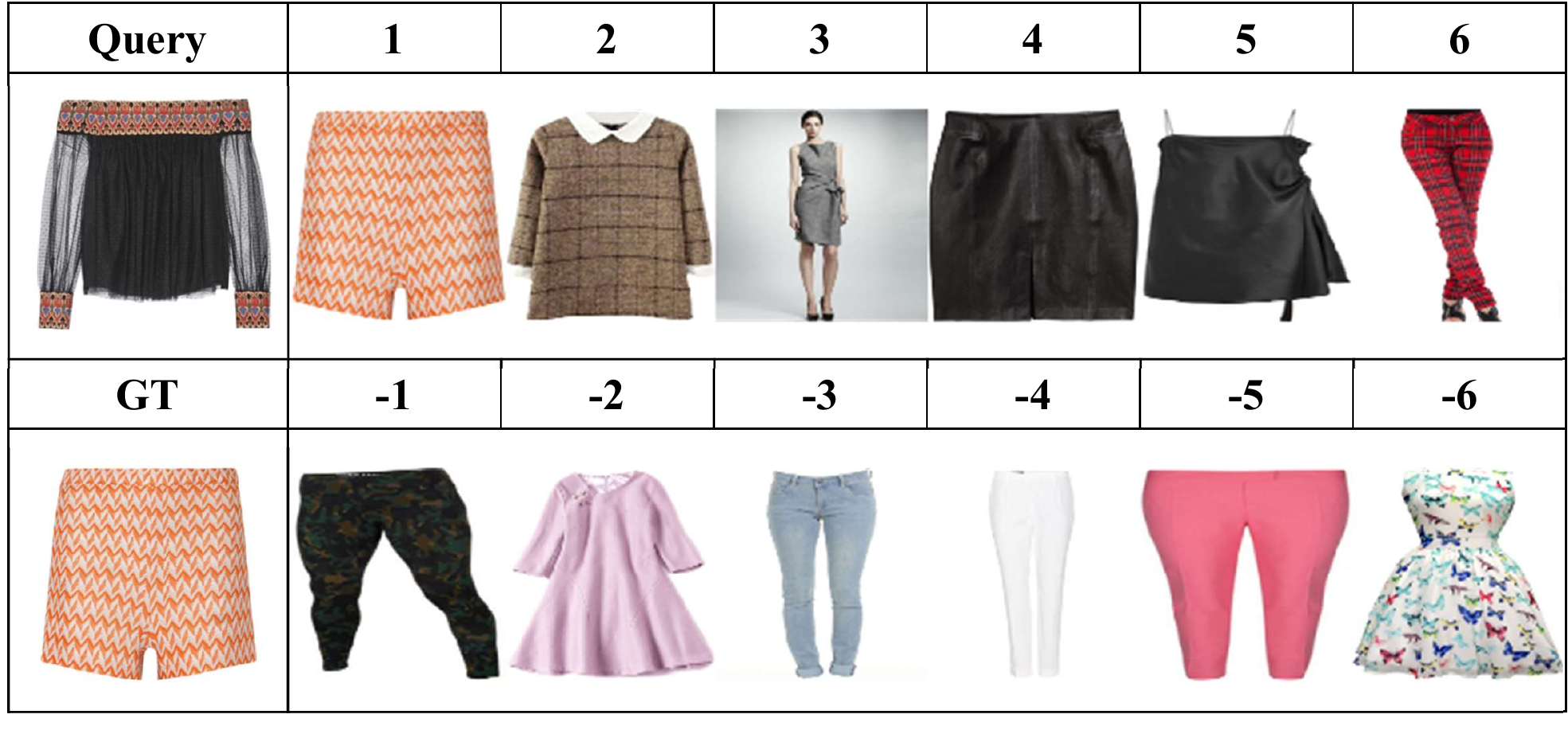}
		\caption{Retrieval results with representation at Layer 4}\label{fig:highlevel}
	\end{subfigure}
	\caption{Example of retrieval results with representations at different layers. We show the query item, ground-truth (GT) item, top 8 results, and last 8 results. Layer 1 is close to the input side and layer 4 is far from the input side.}
	\label{fig:scales}
\end{figure}

\subsection{Prediction Performance}\label{subsec:Prediction}

\noindent\textbf{The prediction performance of MCN.} The results of outfit compatibility prediction AUC and FITB accuracy are shown in \tabref{tab:comparsion}. It can be observed that there is a clear margin between MCN and other baselines. FITB is a harder task because substituting only one part may have little effect on global estimation. BiLSTM performs not well when there is no repeating item of the same type in the outfit, this result is similar to experiments in \cite{Vasileva2018}. Self-Attention computes the relationships between item features as the attention weights, but it does not perform well in the compatibility task because the compatibility cares more about the feature similarities than the feature contents. The pooling method performs better than several complex methods because it is an interaction between all item features. MCN has the same spirit that learns overall compatibility from all pairwise feature similarities.

\begin{figure}[t]
	\centering
	\begin{subfigure}[b]{0.42\textwidth}
		\centering
		\includegraphics[width=\linewidth]{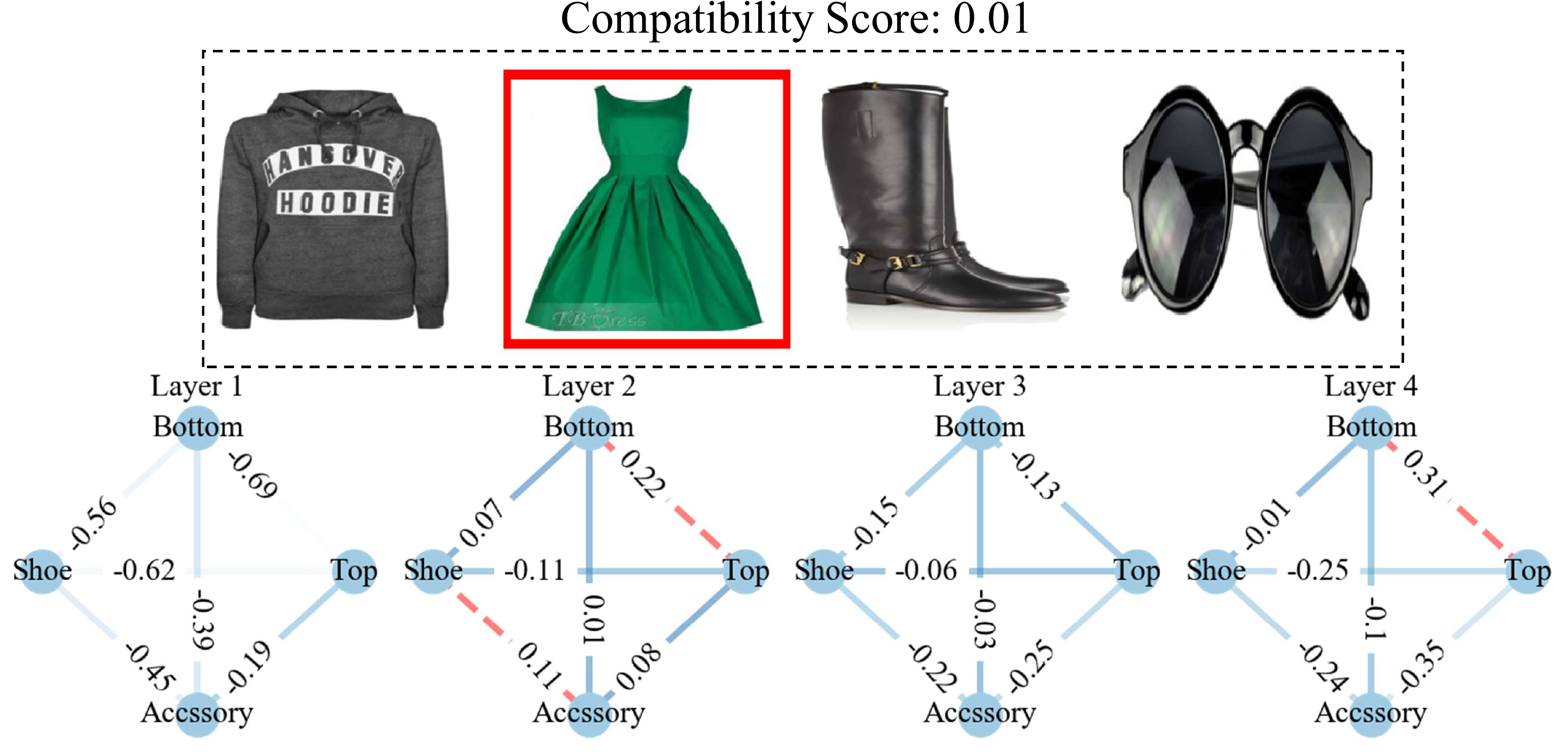}
		\caption{}\label{fig:diagnosis1}
	\end{subfigure}\qquad
	\begin{subfigure}[b]{0.42\textwidth}
		\centering
		\includegraphics[width=\linewidth]{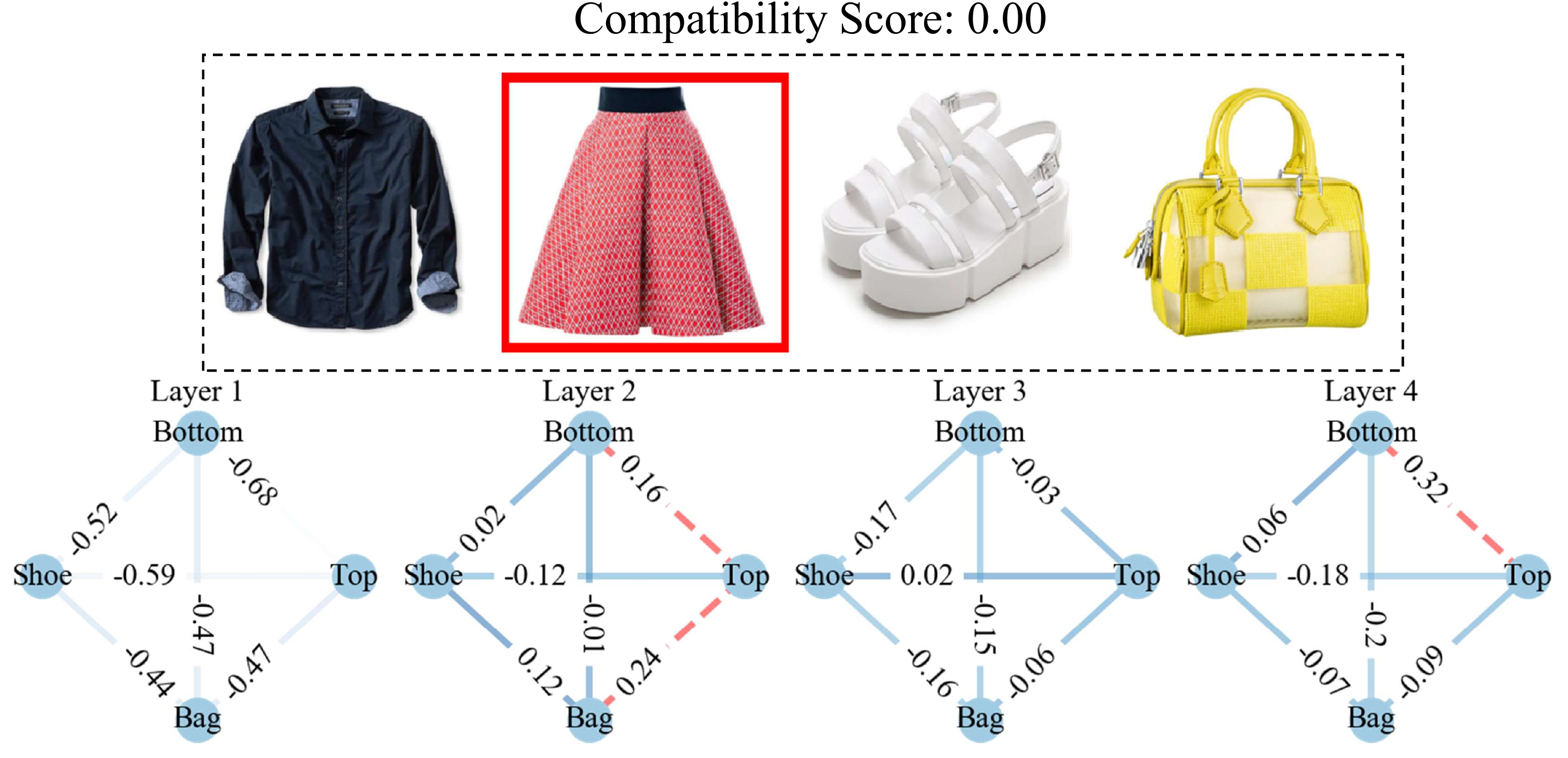}
		\caption{}\label{fig:diagnosis2}
	\end{subfigure}
	\begin{subfigure}[b]{0.42\textwidth}
		\centering
		\includegraphics[width=\linewidth]{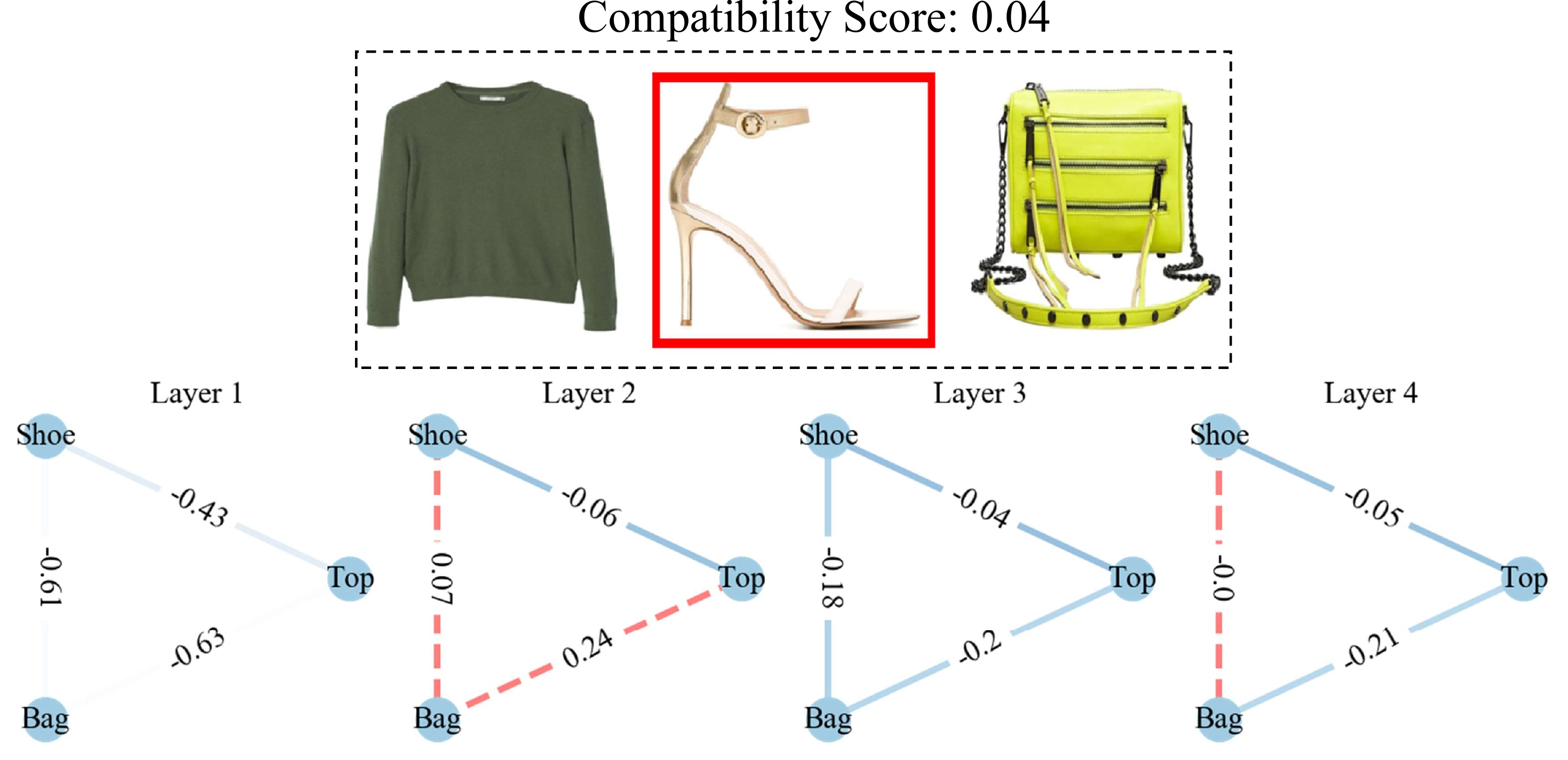}
		\caption{}\label{fig:diagnosis3}
	\end{subfigure}
	\caption{Example of diagnosing the incompatible outfit from different layers.}\label{fig:diagnosis}
\end{figure}

\noindent \textbf{Ablation Study} To analyze the efficiency of each part in the MCN framework, we conduct an ablation study. The results are shown in \tabref{tab:ablation1} where CM stands for the comparison module, VSE is the Visual Semantic Embedding and PE is the Projected Embedding. It can be observed that concatenating item features is the most inefficient method. The comparison module without projection embeddings can already achieve competitive performance and the comparison operation uses fewer parameters because of the removal of MLP with long feature vectors. VSE regularizes the training so there is no additional parameter while evaluation; the parameters of PE have the same magnitude with feature vectors.

\noindent\textbf{What is the relationship between the overall compatibility and pairwise similarities?} As illustrated in \tabref{tab:ablation2}, end-to-end training with comparison module and metric learning method CSN perform similarly. However, CSN has already sampled from the training dataset, so it can not use the same dataset to learn this relationship between the overall compatibility and pairwise similarities. However, the depth of the MLP predictor in MCN can be finetuned to study this problem. We explore 0 to 3 layers and find there is a plateau on 2-layer MLP. It indicates that there is a non-linear relationship between them.

\noindent\textbf{Multi-layered representations benefit the performance.} \figref{fig:scales} is an example of retrieval  with representation at different layers It can be seen that the results retrieved by the low-level features such as layer 1 are similar at color, texture. The results retrieved by high-level features such as layer 4 have similarity in visual style and compatibility. This observation proves the multi-layered representation in MCN ranges from the low level to the high level. \tabref{tab:scales} analyzes the effects of input from different layers. It can be seen that each layer has its contribution to the performance, which supports that the low-level features have an important effect on fashion compatibility.

%\begin{figure}[t]
%	\centering
%	\includegraphics[width=0.9\linewidth]{relation.pdf}
%	\caption{Visualization of output similarties at different layers. Layer 1 is close to the input image side and layer 4 is far from the input image side.}
%	\label{fig:scales}
%\end{figure}

%\begin{table}[h]
%	\begin{center}
%		\caption{Results with different strategies to handle the missing part in outfit of variable length.}
%		\label{tab:blank}
%		\begin{tabular}{|l|c|c|}
%			\hline
%			Method & AUC(\%) & FITB(\%) \\
%			\hline\hline
%			Zero & $86.64  \pm 0.44$ & $56.00 \pm 0.79$ \\
%			Mean & $87.06 \pm 0.52$ & $\mathbf{60.07} \pm 0.32 $ \\
%			Masks & $\mathbf{87.12} \pm 1.11$ & $59.38 \pm 0.68$ \\
%			\hline
%		\end{tabular}
%	\end{center}
%\end{table}

\begin{algorithm}[t]
	\caption{Automatically revise the outfit}\label{alg:revise}
	\begin{algorithmic}[1]
		\Require $X$ - The input images of an outfit containing $N$ items.
		\State $THR = 0.9$ \Comment{Threshold for the compatibility}
		\For {$i$ in ${1 \dots N}$} \Comment{Iterate N times through the outfit}
		\State Find the most problematic item
		\For {each candidate item $x$ with the same type}
		\If {$X$ has better score with the substitution}
		\State Substitute the problematic item with $x$
		\EndIf
		\If {$X$ has compatibility score larger than $THR$}
		\State Return $X$ after substitution
		\EndIf
		\EndFor
		\EndFor
	\end{algorithmic}
\end{algorithm}

%\noindent\textbf{Handle the outfit of variable length.} We explore 3 different strategies to handle the fashion outfit of variable length: (1) Leave the missing item as all zeros which means a black image input. (2) Use a mean image of that type as the padding (3) Take the missing part as all zeros then mask out these zero inputs. We test their performance using comparison module without projected embedding and show their results in \tabref{tab:blank}. It can be observed that the mask strategy and the mean images strategy performs similar (only $0.1\%$ variation), but the zero input strategy performs inferior. The reason may be the similarties between the the zero inputs and others are misleading.

\subsection{Outfit Compatibility Diagnosis}
\textbf{Diagnosis from Different Layers}
The diagnosis process computes the importance of each pairwise similarity and item concerning the incompatibility as described in \secref{sec:diagnosis} We show several diagnosis results in \figref{sec:diagnosis}. All diagnosis values are normalized to let margin between the maximum and minimum as 1. The edges with the 3 greatest values are marked as red dash lines. It can be observed that the diagnosis result is corresponding to human. For instance, in \figref{fig:diagnosis3} the green dress is pointed out mainly because its comparison between the black sweater is incompatible.

\begin{figure}[t]
	\centering
	\begin{subfigure}[b]{0.40\textwidth}
		\centering
		\includegraphics[width=\linewidth]{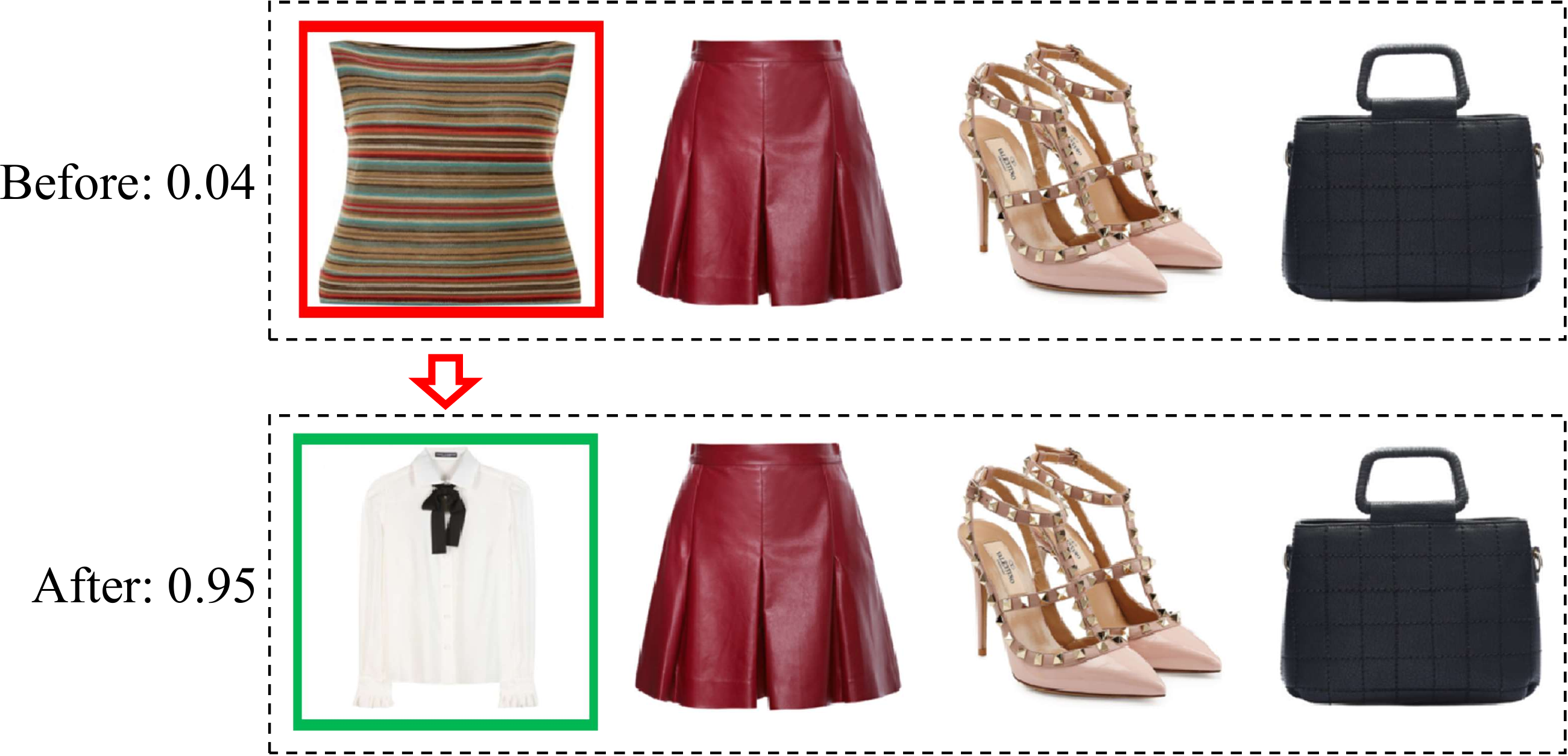}
		\caption{}\label{fig:revise1}
	\end{subfigure}\qquad
	\begin{subfigure}[b]{0.40\textwidth}
		\centering
		\includegraphics[width=\linewidth]{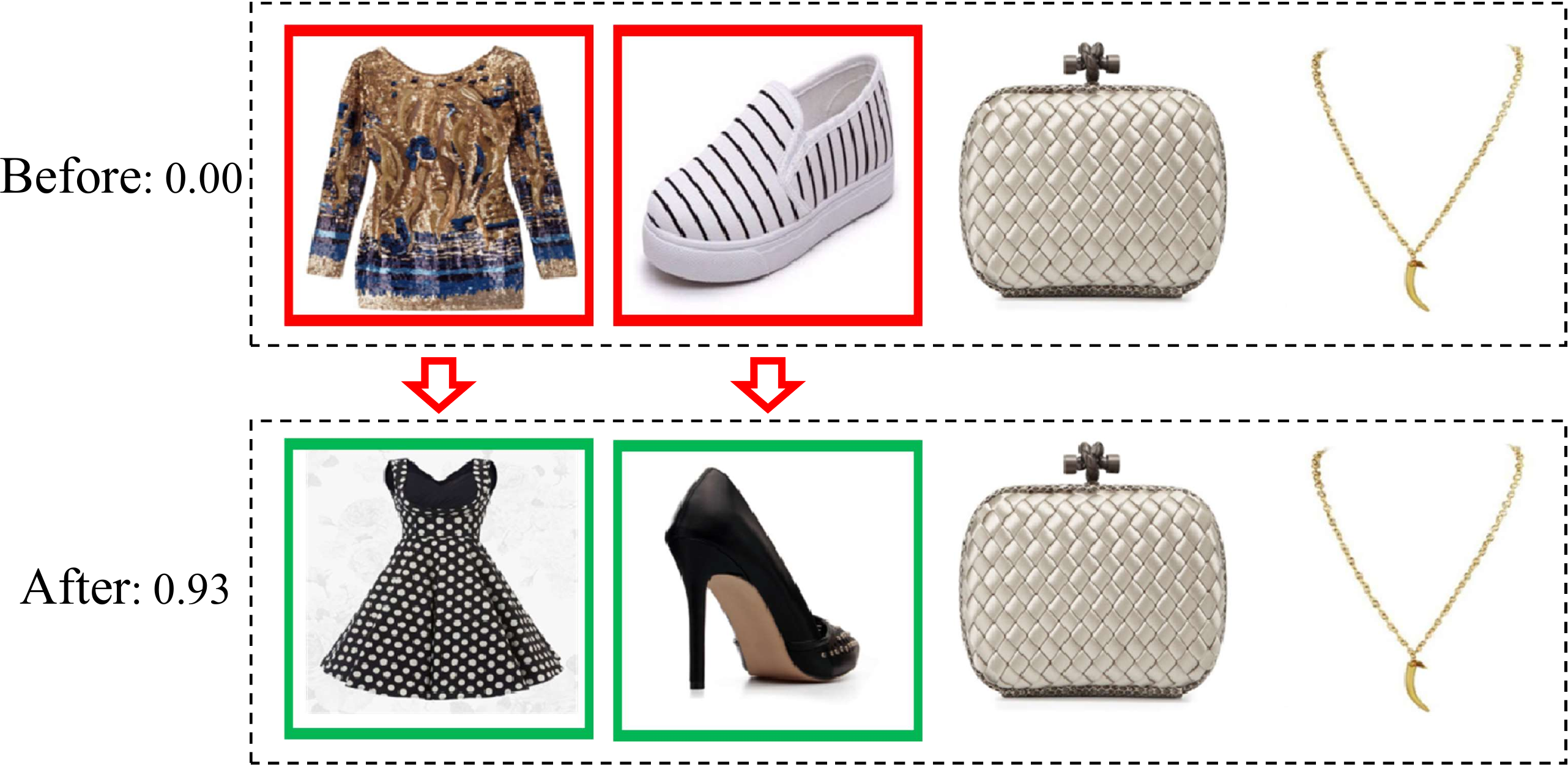}
		\caption{}\label{fig:revise2}
	\end{subfigure}\qquad
	\begin{subfigure}[b]{0.32\textwidth}
		\centering
		\includegraphics[width=\linewidth]{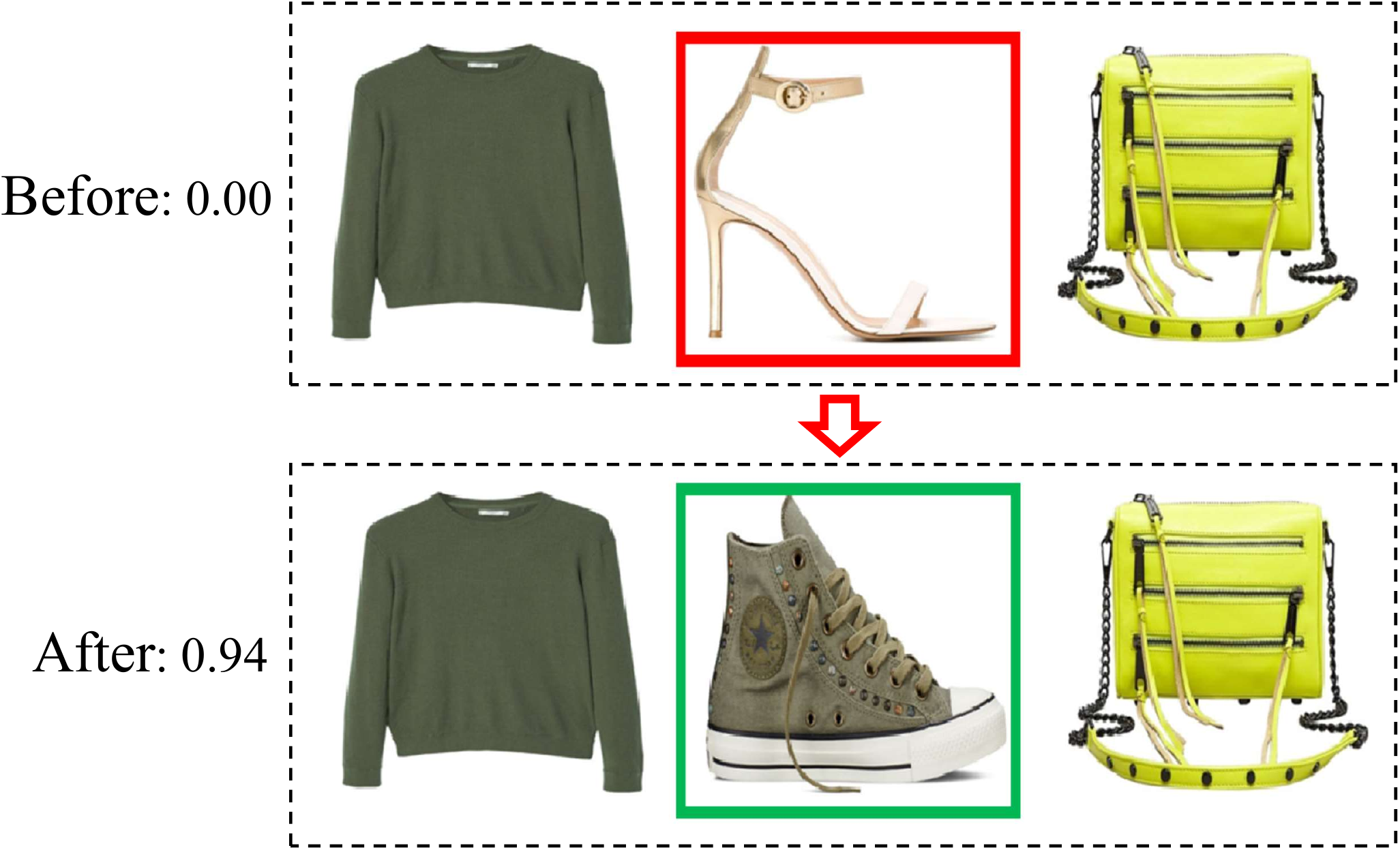}
		\caption{}\label{fig:revise3}
	\end{subfigure}
	\caption{Example of automatically revising the incompatible outfits. }\label{fig:revise}
\end{figure}

\noindent\textbf{Automatically Revise the Outfit}
One benefit of outfit diagnosis is we can revise the outfit based on the diagnosis result, which lets us make very little change to the original composition. We try a simple strategy to substitute the problematic items in the outfit as described in Algorithm \ref{alg:revise}. Several results are shown in \figref{fig:revise}. It can be seen that the revised outfit has a better aesthetic feeling while most items are the same as that in the original outfit.

%The scores on the left represent the predicted compatibility before and after revision.

\section{Conclusion}\label{sec:Conclusion}
In this paper, we introduce an approach to not only predict but also diagnose the fashion outfit. We introduce an end-to-end framework that incorporates the comparison module, multi-layered representation, and visual semantic embedding. The diagnosis process is implemented by first predicting the compatibility score then use the backpropagation gradient to approximate the importance of each compared similarities to the incompatibility. Experiments show that the comparison module is an efficient way to aggregate multiple features for learning outfit compatibility. Representations from different layers also boost the performance which indicates that both low-level features and high-level features have impacts on fashion compatibility. We show that our framework can diagnose the outfit by pointing out the problematic similarities and items, which can be used to interpret the prediction and automatically revise the outfit. For future work, we are planning to explore how to explain different aspects in a better way and quantitively evaluate the diagnosis results.

\begin{acks}
This work is supported by the National Natural Science Foundation of China (Grant No.61572124). Thanks JD AI Research led by Dr. Tao Mei and thanks Wei Zhang for hist insightful discussions.
\end{acks}

%------------------------------------------------------------------------
\clearpage
%
% The next two lines define the bibliography style to be used, and the bibliography file.
\bibliographystyle{ACM-Reference-Format}
\balance
\bibliography{ref}

\end{document}